\documentclass[runningheads]{llncs}

\title{Recent Progress on Graph Partitioning Problems
\\Using Evolutionary Computation}
\author{Hye-Jin Kim \and Yong-Hyuk Kim} 
\institute{Department of Computer Science, Kwangwoon University \\20 Kwangwoon-ro, Nowon-gu, Seoul 01897, Republic of Korea\\
\email{email: ovoa22@gmail.com, yhdfly@kw.ac.kr}}
\begin{document}
\maketitle

\begin{abstract}
The graph partitioning problem (GPP) is a representative combinatorial optimization problem which is NP-hard. Currently, various approaches to solve GPP have been introduced. Among these, the GPP solution using evolutionary computation (EC) is an effective approach. There has not been any survey on the research applying EC to GPP since 2011. In this survey, we introduce various attempts to apply EC to GPP made in the recent seven years. 
\end{abstract}

\section{Introduction}
The graph partitioning problem (GPP) is an NP-hard combinatorial optimization problem. In other words, it is a problem with the same subset size as the minimal number of cut sizes when dividing a graph into two or more subsets. The $k$-way graph partitioning problem is a problem to partition a graph into $k$ subsets. When $k$ = 2, it is referred as graph bisection or bi-partitioning. As obtaining accurate optima of GPP is rather challenging, we can obtain approximate solutions with approaches such as the Kernighan-Lin algorithm \cite{kernighan70}, Fiduccia-Mattheyses algorithm \cite{fiduccia88}, and multi-level method. Furthermore, a heuristic approach is often adopted depending on the nature of problem. Kim \textit{et al.} \cite{kim11genetic} first reviewed researches focusing on genetic algorithms. After this, there have been much studies on GPP using evolutionary computation, but there has been no research surveying these studies. Recently, Bulu{\c{c}} \textit{et al.} \cite{bulucc16} conducted a survey to examine recent trends in graph partitioning approaches, but it was a comprehensive survey research focusing on GPP. In this research, we organized a stream of research over the last seven years on evolutionary computation to solve GPP. This paper is organized as follows. In section 2, research related to the encoding of genetic algorithm (GA) is covered. In Section 3, local optimization used in the multi-level heuristic, and the hybrid method are introduced. In Section 4, other approaches are covered. A summary of the paper is presented in Section 5.

\section{Encoding}
Yoon and Kim \cite{yoon14} studied heuristics for vertex ordering and an approach to partition vertex ordering into clusters. This approach was applied to integrate the reordering of GA and identification of clustered structures in graphs. Kim \textit{et al.} \cite{seo2015edge} introduced edge-set encoding based on spanning trees. Unlike the existing vertex-based encoding, edge subsets derived from the spanning tree correspond to the genes in the encoding. This encoding has the advantage of presenting feasible solutions only, so there is no need to apply a repair process. 

\section{Local Optimization}
\subsection{Multi-level Heuristics}
Sanders and Schulz \cite{sanders2012distributed,sanders2012} used multi-level graph partitioners to solve GPP. This partitioner provides effective crossover and mutation operators. The crossover and mutation operators were incorporated with an extendible communication protocol, which allows many inputs to be processed in a short period of time. Pope \textit{et al.} \cite{pope16} used genetic programming to develop the multi-level graph partition method adjusted for a specific application program. Using this, they demonstrated the results using evolved partitioners in traditional random graph models, as well as in a real-world computer network dataset. Andre \textit{et al.} \cite{andre17} focused on the quality of solutions when solving problems and developed a multi-level memetic algorithm. To develop this algorithm, effective multi-level recombination and mutation processes were developed. Biedermann \textit{et al.} \cite{biedermann18} proposed a general memetic algorithm to solve a graph clustering problem. In this algorithm, natural recombination operators using ensemble clustering and multi-level techniques were adopted.

\subsection{Local Search Algorithm in Hybrid GAs}
Lee \textit{et al.} \cite{lee12} presented a circuit bipartitioning algorithm. The key idea of this algorithm is to embed local search heuristics into a quantum-inspired evolutionary algorithm (QEA). The Fiduccia-Mattheyses algorithm \cite{fiduccia88} was used for local optimization and operators were revised to apply QEA. Hwang \textit{et al.} \cite{hwang15} demonstrated heuristic distinguishing clusters in the population of locally optimized random partitions. This becomes the seed of the memetic algorithm (MA), and once MA is executed, beneficial cluster moves can be performed. Kim \cite{Kim16an} presented an incremental GA to solve the GPP. By proposing graph expansion methodologies and vertex ordering schemes, Kim defined an appropriate sequence of subproblems. Kim \textit{et al.} \cite{kim18opt} proposed a simulated catalytic reaction method to explore optimization problems. With the catalytic reactions applied to optimization problems, one can figure out the hidden information, and to solve GPP, an applied framework is also proposed.

\section{Other Approaches}
First, among non-traditional GAs, Wang \textit{et al.} \cite{wang14} proposed an algorithm that improved QEA through linkage learning. The existing QEAs synchronize individuals to prevent QEA from finding multiple optima, while this algorithm overcame this shortcoming. In addition, the niching approach replaced migration to increase the convergence speed, avoid genetic drift, and locate all global optima.
There is also research adopting parallel GAs. Wang \textit{et al.} \cite{wang2015genetic} developed a nonlinear mixed integer programming model. In this problem, they devised a parallel GA to solve large-scale, real-world instances.
In GPP research using a particle swarm optimization (PSO) algorithm, Li \textit{et al.} \cite{li16} proposed a link density-based multi-objective optimization model to solve a community detection problem. Also, a multi-objective PSO algorithm was presented to solve this model. By using each single run of the proposed algorithm, one can generate a Pareto solution set, and a network community structure can be presented as the set is uniformly distributed. 
There are two studies that solved GPP using ant colony optimization (ACO). Sachdeva and Kaur \cite{sachdeva15}  partitioned an application program to minimize energy consumption in an applied GPP problem. This study proposed an ACO-based approach to deploy software in the mobile cloud. Liu \textit{et al.} \cite{liu16} used the MapReduce programming model and proposed ant brood clustering with intelligent ants that cluster financial time series data. The algorithm used an alternate number of mappers in each iteration for data parallelism and indirect communication between mappers. 
Research has also been conducted on the harmony search algorithm. Kim \textit{et al.} \cite{kim18acomparison} was the first to use a harmony search algorithm to solve a max-cut problem. Compared with GA, harmony search algorithm is better in that good individuals can be generated at each iteration. After considering all existing individuals, fewer parameters should be determined compared with those of GA, which is seen as a major advantage of this algorithm. 
Lastly, Zadeh and Kobti \cite{zadeh15} developed a multi-population cultural algorithm using knowledge-based evolutionary computation to solve GPP-applied problems. This algorithm guides search direction as knowledge is derived from the network and the optimal solution is found accordingly.

\section{Summary}
Some applications of GPP are circuit placement and community detection, and they occur in varying fields. Research on heuristic approaches to solve GPP has been continuously conducted. The evolutionary approach is also viewed as an effective approach to expect fair outcomes in GPP. In Table 1, GPP studies using evolutionary computation (EC) shown in this research are summarized. Many researchers mentioned in the study used various EC approaches to solve GPP. Moreover, they achieved competitive results to demonstrate the effects of algorithms built upon the results. This research summarized and organized various EC-based approaches, such as encoding and local optimization, applied to GPP. The research is expected to contribute to future research using EC approaches to solve GPP.

\begin{table*}\begin{footnotesize}
  \caption{List of papers using evolutionary approach for GPP}
  \label{tab:commands}
  \begin{tabular}{|c|c|c|c|c|}
  \hline
  
  \textbf{Author(s)} & \textbf{Brief descriptions} & \textbf{Year} & \textbf{Section} & \textbf{Reference}\\
  \hline
  \hline
  Kim \textit{et al.} & survey of EC approaches & 2011 & 1 & \cite{kim11genetic}\\ 
  \hline
  Sanders and Schulz & distributed evolutionary algorithm, multi-level heuristic & 2011 & 3.1 & \cite{sanders2012distributed,sanders2012}\\
  \hline
  Lee \textit{et al.} & memetic quantum-inspired evolutionary algorithm & 2012 & 3.2 & \cite{lee12}\\
  \hline
  Yoon and Kim & vertex ordering & 2014 & 2 & \cite{yoon14}\\
  \hline
  Wang \textit{et al.} & quantum-inspired evolutionary algorithm & 2014 & 4 & \cite{wang14}\\
  \hline
  Seo \textit{et al.} & edge-based encoding & 2015 & 2 & \cite{seo2015edge}\\
  \hline
  Hwang \textit{et al.} & cluster-handling heuristic & 2015 & 3.2 & \cite{hwang15}\\
  \hline
  Wang \textit{et al.} & parallel GA & 2015 & 4 & \cite{wang2015genetic}\\
  \hline
  Sachdeva and Kaur & ant colony optimization & 2015 & 4 & \cite{sachdeva15}\\
  \hline
  Zadeh and Kobti & multi-population cultural algorithm & 2015 & 4 & \cite{zadeh15}\\
  \hline
  Bulu{\c{c}} \textit{et al.} & survey of general approaches & 2016 & 1 & \cite{bulucc16}\\
  \hline
  Kim & hybrid GA & 2016 & 3.2 & \cite{Kim16an}\\
  \hline
  Pope \textit{et al.} & multi-level partitioning & 2016 & 3.1 & \cite{pope16}\\
  \hline
  Li \textit{et al.} & particle swarm optimization & 2016 & 4 & \cite{li16}\\
  \hline
  Liu \textit{et al.} & ant brood clustering & 2016 & 4 & \cite{liu16}\\
  \hline
  Andre \textit{et al.} & multi-level memetic algorithm & 2017 & 3.1 & \cite{andre17}\\
  \hline
  Biedermann  \textit{et al.} & memetic algorithm & 2018 & 3.1 & \cite{biedermann18}\\
  \hline
  Kim and Kang & simulated catalytic reaction method, local search & 2018 & 3.2 & \cite{kim18opt}\\
  \hline
  Kim \textit{et al.} & harmony search & 2018 & 4 & \cite{kim18acomparison}\\
  \hline
  \end{tabular}
  \end{footnotesize}
\end{table*}


\bibliographystyle{acm}
\bibliography{bibliography} 

\end{document}